\PassOptionsToPackage{table}{xcolor}
  \documentclass[runningheads]{llncs}

\usepackage[mobile]{eccv}
\usepackage{wrapfig}
\usepackage{CJKutf8}
\usepackage{capt-of}

\usepackage{eccvabbrv}

\usepackage{graphicx}
\usepackage{booktabs}
\usepackage{rotating}

\usepackage[accsupp]{axessibility}  

\usepackage{hyperref}

\usepackage{orcidlink}
\usepackage{amssymb}   
\usepackage{pifont}    

\begin{document}

\title{Question-Guided Evidence Acquisition for Multimodal Visual Question Answering}

\titlerunning{Q-Guide}

\author{Alin-Ionut Popa}

\authorrunning{A.-I. Popa}

\institute{
Amazon Inc. \\
\email{popaaln@amazon.com}}

\maketitle

\begin{center}
\begin{minipage}{\textwidth}
\centering
\includegraphics[width=0.95\textwidth]{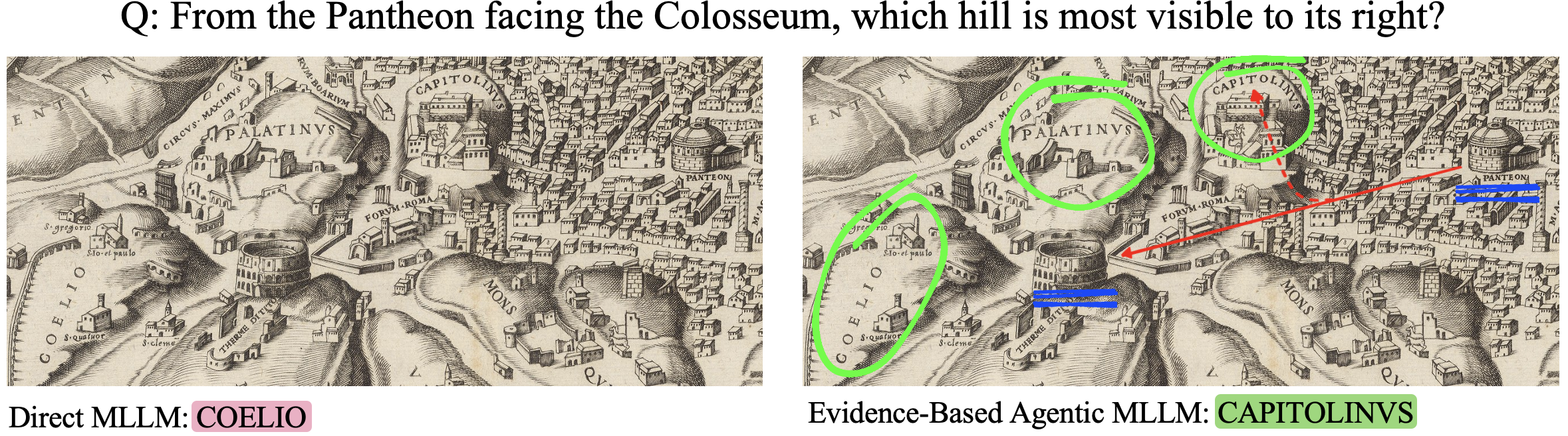}
\captionof{figure}{\textbf{From uniform perception to question-conditioned evidence recovery.} Direct MLLM inference (\textit{Left}) processes the document uniformly and answers from incomplete perception. Q-Guide (\textit{Right}) directs perception toward the question---locating landmarks, reading targeted labels, and verifying spatial relationships---before committing to an answer.}
\label{fig:teaser}
\end{minipage}
\end{center}


\begin{abstract}
Multimodal LLMs can see a document, but they often can't read it reliably. Small text, tables, visual cues, and topological elements still trip them up under direct visual inference, even when the page is already sitting in the model's context. Most document-VQA systems treat perception as fixed: they encode the page once and answer from whatever the model extracted in that single fast pass. We think document VQA needs slower, more deliberate perception: instead of committing to one fixed encoding, the model should spend a little extra compute at inference time working out what to look at next, and only then answer. We build this into \textbf{Q-Guide}, a small agent that reads a question, works out what evidence it is still missing, and calls targeted tool(s) to recover it---reading text where text is needed, zooming in where detail is needed, or grounding a region where position matters. On DocVQA2026 and M109NC (a Manga109-based character-naming task), Q-Guide outperforms both direct prompting and recent multi-agent document systems ($65.0\%$ vs.\ $40.0\%$ on DocVQA2026, $32.4\%$ vs.\ $24.4\%$ on M109NC), and the improvement holds across three Claude backbones (Opus 4.6, Sonnet 4.6, and Opus 4.5). We find that accuracy scales with the perception budget---most of the gain appears within two to three deliberate rounds---and that the gain comes from directing perception to the right place, not from complex control logic: adding planners, routers, or multiple collaborating agents does not help.
\end{abstract}

\section{Introduction}
\label{sec:intro}

Multimodal large language models have become very good at looking at documents, grounding regions, and reasoning across modalities~\cite{yin2024mllmsurvey,ding2026vrdusurvey,bai2025qwen25vl,qin2025covt}, building on steady progress in visual object representation and detection~\cite{Leotescu_2025_WACV,10031080}. But there is still a gap between seeing a document and actually reading it, and that gap grows as documents get visually denser. A model can have a page in its context window and still misread a dimension on an engineering drawing, miss which table cell is highlighted, or overlook a small label on a map (Figure~\ref{fig:teaser}). In our experiments this was the most common failure: the model could reason fine, but it did not reliably perceive the specific piece of evidence the question depended on.

\begin{figure*}[t]
    \centering
    \includegraphics[width=0.95\textwidth]{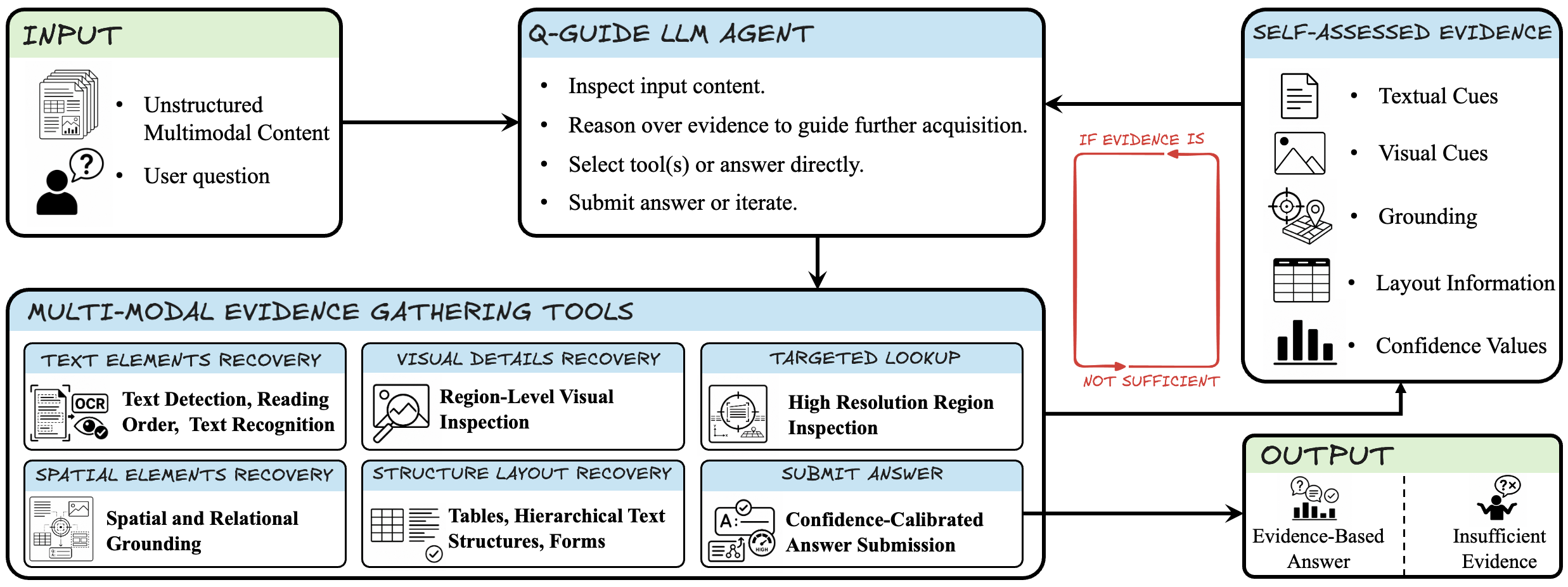}
    \caption{\textbf{Overview of Q-Guide.} Given unstructured multimodal content and a user question, Q-Guide makes perception question-conditioned: the agent inspects the document, reasons about what evidence is still missing, selects perceptual recovery tools to fill the gap, and updates its evidence state. The loop continues until the recovered evidence supports an answer.}
    \label{fig:detailed_overview}
\end{figure*}

One reason is that most document-VQA pipelines perceive the document in a single fast pass, the same way regardless of the question~\cite{docvqa2026dataset,hu2024docowl2,liu2024textmonkey,bai2025qwen25vl}. The page is encoded once and the model answers from whatever it extracted---a System-1 style that commits before it has really looked. This works when the evidence is large and central, but visually rich documents are not like that~\cite{mo2025doccob,yu2025bboxdocvqa,wang2025marten,luan2024textcot}: the evidence for one question might be a value printed in red, for the next a route on a map or a name in a speech bubble, each needing a different way of being read.

Our idea is to make perception slow and deliberate, and to condition it on the question~\cite{yin2025toolvqa,luan2024textcot,hu2024visualsketchpad,mohammadshirazi2025arial,wang2026agenticocr}. Rather than committing after one pass, the model asks what this particular question requires it to perceive, spends test-time compute recovering exactly that, and repeats until it has enough evidence to answer---a System-2 loop over perception rather than over language alone. We build this into a compact agent we call \textbf{Q-Guide}.

\textbf{Contributions.} (i) We cast document VQA as test-time perceptual reasoning: a question-conditioned loop that spends extra compute on \emph{what to perceive} beats heavier reasoning orchestration (planning agents, routing, and multi-agent collaboration) on the same backbone~\cite{jiang2024multiagentvqa,sun2025docagent,han2025mdocagent,zhang2025which}, and accuracy scales with the perception budget until it saturates around two to three rounds. (ii) The same design transfers to very different visual domains, from document pages to manga, without any structural changes. (iii) The gains are not specific to one model: they hold across Claude Opus 4.6, Sonnet 4.6, and Opus 4.5.

We evaluate on two benchmarks that stress different things. DocVQA2026~\cite{docvqa2026dataset} spans eight document categories with reasoning over up to 181 pages, and tests breadth across layouts, tables, and long documents. M109NC, a character-naming task we build from Manga109~\cite{multimedia_aizawa_2020,baek2026mangav26,vivoli2024comix}, tests one hard axis instead: matching a character's identity across pages from Japanese dialogue and visual appearance. On both, Q-Guide beats direct prompting and recent agentic baselines~\cite{mohammadshirazi2025arial,sun2025docagent,han2025mdocagent}. Since these baselines were originally reported on different MLLM backbones, we re-implement each on our Claude backbones---keeping each method's own tools and control logic---so any difference reflects the evidence-acquisition strategy, not the underlying model. Ablations then confirm the gains come from directing perception toward question-relevant evidence; adding orchestration layers does not help and sometimes hurts~\cite{zhang2025which}.

\section{Related Work}
\label{sec:related}

The gap we described in the introduction---models that see a page but do not reliably read the evidence a question needs---has been approached from two directions: model-side, training or adapting MLLMs to perceive text-rich documents better, including long-range sequence encoders for document structure~\cite{leotescu2024bidirectionallongrangeparsersequential,sandu-etal-2022-large}; and system-side, keeping a general-purpose MLLM and giving it external perception it can call on. Q-Guide is system-side: the MLLM stays as-is but gains question-conditioned recovery tools that surface evidence it cannot reliably extract from pixels alone~\cite{ding2026vrdusurvey,hu2024docowl2,liu2024textmonkey,cho2024m3docrag,yu2024visrag,docvqa2026dataset,multimedia_aizawa_2020,baek2026mangav26}.

\textbf{OCR-free and document-specialized MLLMs.}
OCR-free and document-specialized MLLMs improve the model-side representation of text-rich documents~\cite{hu2024docowl2,liu2024textmonkey,liao2024doclayllm,cui2025paddleocrvl,bai2025qwen25vl,rodriguez2024bigdocs,qin2025covt}, building on a longer line of specialized text-spotting in document images~\cite{krubinski2024watermarktextpatternspotting}, internalizing document perception through architecture, training data, or visual encoding. Q-Guide goes the other way, exposing question-conditioned recovery tools the model can invoke on demand~\cite{amazontextract,suryaocr}.

\textbf{Tool-augmented and agentic VQA.}
There is growing interest in moving VQA from passive answer generation toward active evidence gathering~\cite{yin2025toolvqa,luan2024textcot,liu2024chainofspot,hu2024visualsketchpad}. Several recent methods let vision-language models call external perception modules, break down visual queries, and iteratively collect information before committing to an answer~\cite{yin2025toolvqa,jiang2024multiagentvqa,reddy2025orion}. In visually rich document understanding, closely related systems such as ARIAL~\cite{mohammadshirazi2025arial} and AgenticOCR~\cite{wang2026agenticocr} treat document QA as an iterative process of locating and parsing evidence rather than relying on a fixed OCR transcript. Q-Guide shares this motivation but asks a more pointed question: which perceptual recovery actions actually help when conditioned on the question, and which ones just add overhead?

\textbf{Agentic orchestration for document question answering.}
Recent document QA systems explore increasingly structured forms of orchestration, including planning agents, multi-agent collaboration, retrieval modules, answer grounding, and iterative self-correction~\cite{mohammadshirazi2025arial,sun2025docagent,han2025mdocagent,wen2026ocragent,cho2024m3docrag,yu2024visrag,faysse2024colpali,tanaka2025vdocrag,dong2025mmdocrag,jain2025simpledoc,lopez2025enhancingdocvqa,zhang2026pivotbridgingplanningexecution}. These systems show that decomposing document understanding into smaller actions can improve interpretability and robustness, especially for long or visually complex documents~\cite{mo2025doccob,yu2025bboxdocvqa,wang2025marten,indrehus2026coexvqa}. The downside is that heavier orchestration brings its own failure modes: planner errors, noisy intermediate summaries, repeated reflection, and context that ends up distracting the reasoning model~\cite{zhang2025which}. Q-Guide tests this trade-off head-on: rather than deeper orchestration, it uses a compact single-loop agent that directs perception toward question-relevant evidence and stops when the recovered modalities support an answer.

\section{Q-Guide}
\label{sec:approach}

The central idea is to make perception adaptive to the question. Rather than applying a fixed processing pipeline, the model iteratively identifies what evidence is missing and recovers it through targeted tools. Figure~\ref{fig:detailed_overview} gives an overview. The loop is short: the unstructured multimodal content and the user question come in, the Q-Guide agent inspects the document and picks question-conditioned recovery actions, the recovery tools extract the missing evidence, and the agent judges whether it now has enough to answer or needs another round.

\subsection{Preliminaries and Formulation}
\label{sec:formulation}

Let $\mathcal{D} = \{p_1, \ldots, p_N\}$ be a document of $N$ page images and $q$ the user question. Q-Guide answers $q$---or returns \texttt{Unknown} when the evidence is insufficient---by iteratively gathering evidence and testing whether it suffices, an interaction defined by the tuple $(\mathcal{E}, \mathcal{A}, \mathcal{T}, \phi)$.

\textbf{Evidence state.} At turn $t$ the agent holds an evidence state $e_t = (q, \mathcal{D}, O_t) \in \mathcal{E}$, where $O_t = (o_1, \ldots, o_{k_t})$ is the sequence of observations gathered so far, starting empty. Each observation $o_i = (a_i, r_i, y_i)$ records the tool $a_i$, the normalized region $r_i = (l,t,r,b) \in [0,1]^4$ it read from, and the returned content $y_i$. Attaching the source region $r_i$ to every observation ties each answer to the page location it was read from rather than to parametric priors.

\textbf{Actions and transition.} The action space $\mathcal{A} = \mathcal{A}_{\text{tool}} \cup \{\texttt{submit}\}$ holds the five recovery tools (Section~\ref{sec:tools}) and the answer submission. A single policy $\pi$---the same LLM at every turn---proposes actions $\pi(e_t) \subseteq \mathcal{A}$ from the current evidence; the termination rule below arbitrates this proposal into the executed set $\mathbf{a}_t$, whose tool actions append their observations to the state:
\begin{equation}
    e_{t+1} = \mathcal{T}(e_t, \mathbf{a}_t)
    = \Big(q,\; \mathcal{D},\; O_t \oplus \!\!\bigoplus_{a \in \mathbf{a}_t \cap \mathcal{A}_{\text{tool}}}\!\! \texttt{execute}(a)\Big),
    \label{eq:transition}
\end{equation}
where $\oplus$ concatenates the new observations onto $O_t$ and \texttt{submit} adds none. Because the transition only ever \emph{appends} to $O_t$ and the agent places just $P_{\text{ctx}}\!\ll\!N$ pages in context (of the $N$ total), the interaction cost scales with the number of rounds and the evidence gathered, not with the document length $N$---a property we quantify in Section~\ref{sec:failure_discussion}.

\textbf{Sufficiency and termination.} Whether to gather more or answer is governed by the sufficiency predicate $\phi : \mathcal{E} \to \{0,1\}$---the agent's judgment of whether the evidence suffices---which we make explicit. On submission the policy emits a self-reported confidence level $\sigma(e_t)\in\mathcal{L}=\{\textsc{low}\prec\textsc{med}\prec\textsc{high}\}$. The gate commits either when this confidence is at or above a threshold $\tau$, or---for a less-confident submission---when a lightweight check finds the answer explicitly supported by the gathered evidence, $\textsc{ver}(e_t)=1$:
\begin{equation}
    \phi(e_t) \;=\; \mathbb{1}\!\left[\,\text{policy submits}\;\wedge\;\big(\sigma(e_t)\succeq\tau \;\vee\; \textsc{ver}(e_t)\big)\,\right],
    \qquad \tau=\textsc{med}\ \text{(default)},
    \label{eq:phi}
\end{equation}
with $\mathbb{1}[\cdot]$ the indicator ($1$ iff its argument holds), $\succeq$ the order on $\mathcal{L}$, and $\textsc{ver}(e_t)=1$ iff the candidate answer is explicitly supported by some observation $o_i\in O_t$. In words, high- and medium-confidence answers pass directly, while a less-confident answer commits only if verification backs it and otherwise abstains---a test applied by the policy itself, not a separate critic module. Each turn then resolves as:
\begin{equation}
    \mathbf{a}_t =
    \begin{cases}
        \{\texttt{submit}(e_t)\} & \text{if } \phi(e_t)=1, \\
        \{\texttt{submit\_unknown}(e_t)\} & \text{if } \phi(e_t)=0 \text{ and } \big(\text{policy submits} \text{ or } t \geq T_{\max}\big), \\
        \pi(e_t) & \text{otherwise (gather more)}.
    \end{cases}
    \label{eq:termination}
\end{equation}
The second case is where verification does its work: a below-threshold submission that the evidence does not explicitly support ($\phi=0$ with the policy nonetheless submitting) abstains to \texttt{Unknown} rather than emitting an unsupported answer; the same abstention applies once the budget $T_{\max}$ is spent. Because one model both selects actions and evaluates $\phi$, Q-Guide needs no separate planner, router, or critic; it simply reasons over the accumulated evidence and decides what to perceive next.

\begin{figure}[!htbp]
\centering
\includegraphics[width=0.8\textwidth]{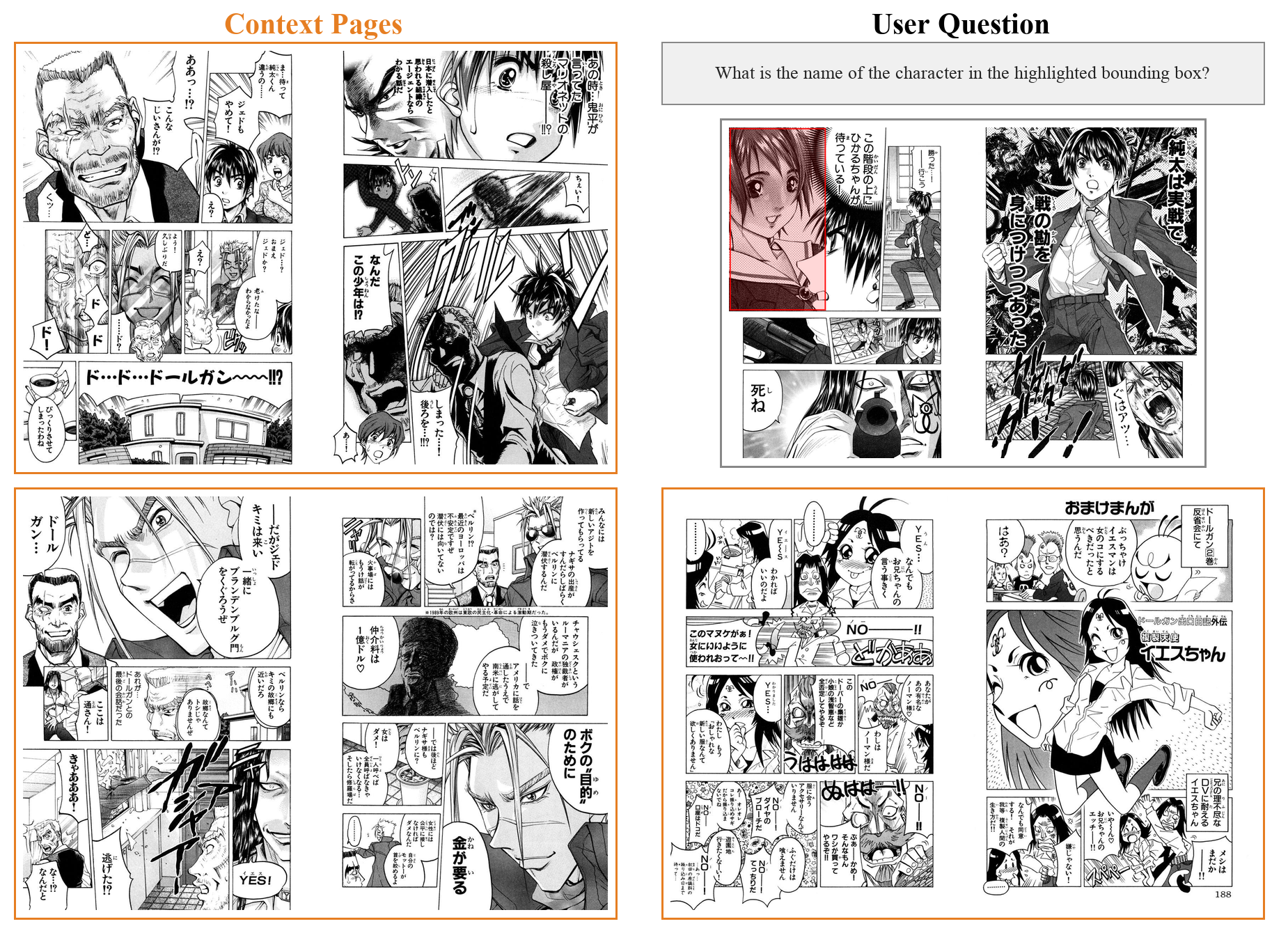}
\vspace{-2mm}
\caption{\textbf{M109NC task example.} For visualization, we show 3 of the 6 context pages (orange border) that reference the query character; the query page highlights the target (red overlay). The task requires cross-page reasoning: reading Japanese dialogue, associating names with visual appearances, and matching the target's identity across scenes.}
\label{fig:manga109_sample}
\end{figure}

\subsection{Architecture}
\label{sec:architecture}

Q-Guide instantiates the loop of Section~\ref{sec:formulation} with three parts. First, the \textbf{Q-Guide agent} (an MLLM) is the single policy $\pi$: it receives the unstructured multimodal content and the question, and at each turn inspects the evidence state $e_t$, reasons about what perceptual evidence is still missing, and selects one or more recovery tools to fill the gap. Multiple operators can be invoked in a single turn for complementary evidence. The instructions are intentionally minimal---short operator descriptions and answer-format guidance, with no explicit planning, category routing, or advisory hints.
Second, the selected \textbf{question-conditioned recovery tools} extract information that is difficult to perceive from the raw image alone---text, visual detail, spatial grounding, layout structure, and targeted region answers---each addressing a perceptual gap conditioned on the current question. Every call returns a grounded observation $o_i=(a_i,r_i,y_i)$ that the transition $\mathcal{T}$ appends to the evidence state.
Third, Q-Guide performs \textbf{self-assessment}: the same policy that directs perception also evaluates the sufficiency predicate $\phi$ over the accumulated observations, looping back for more evidence while $\phi=0$ and submitting once $\phi=1$. Because the state grows one deliberate perception step at a time, this loop is where the model's test-time compute is spent.

Finally, the \textbf{output} follows Eq.~\ref{eq:termination}: an evidence-based answer once $\phi(e_t)=1$, or \texttt{Unknown} when the agent exhausts its budget or a low-confidence answer fails verification.

\subsection{Question-Conditioned Recovery Tools}
\label{sec:tools}
The model has direct access to the full document, yet seeing is not reading: small text, dense tables, colored cells, spatial arrangements, and stylized fonts all degrade under direct visual inference, so active recovery is needed even though the pixels are already in context. Each tool below targets one such gap, and the agent decides which to fill next based on what the question demands and what it has gathered so far.

Q-Guide exposes five evidence-recovery tools and one submission action, matching the tool groups in Figure~\ref{fig:detailed_overview}. Each tool operates on a document page or on a region specified by normalized coordinates $(l,t,r,b) \in [0,1]^4$. This allows the agent to move from a global view of the document to targeted local evidence.

\textbf{Text elements recovery.}
This tool recovers text from a selected page or region. It supports text detection, reading-order recovery, and text recognition. It is useful for small text, scattered labels, dimensions, captions, form fields, and dense document regions.

\textbf{Visual details recovery.}
This tool returns a high-resolution crop of a selected region. It does not perform OCR; instead, it gives the LLM a clearer view of the raw pixels. It is useful for color, shape, layout, relative position, chart appearance, and other details that may be lost in text extraction.

\textbf{Targeted lookup.}
This tool asks a natural-language question over a selected region and returns a short answer with confidence. It is useful when the agent has already localized the relevant area and needs a precise value, such as a total, width, label, or table entry.

\textbf{Spatial elements recovery.}
This tool provides grounding capabilities that anchor text and visual elements in the input coordinate space, recovering word-level positions, color distributions, and text along geometric corridors. It is useful for highlighted values, map-like layouts, or text arranged along a path.

\begin{wraptable}{r}{0.52\textwidth}
\vspace{-10pt}
\centering
\small
\setlength{\tabcolsep}{4pt}
\renewcommand{\arraystretch}{1.12}
\caption{\textbf{Tool importance on DocVQA2026.} Leave-one-out over the five recovery tools (T: Text, V: Visual Inspection, Q: Targeted Query, S: Structure, P: Spatial), reporting accuracy and the drop when each is removed. Visual Inspection and Targeted Query are the most impactful; analyzed in Section~\ref{sec:ablation}.}
\label{tab:tool_ablation}
\scalebox{0.86}{%
\begin{tabular}{@{}l ccccc cc@{}}
\toprule
\textbf{Configuration}
& \textbf{T} & \textbf{V} & \textbf{Q} & \textbf{S} & \textbf{P}
& \textbf{Acc.} & \textbf{Drop} \\
\midrule
Q-Guide (full)
& $\checkmark$ & $\checkmark$ & $\checkmark$ & $\checkmark$ & $\checkmark$
& $65.0\%$ & \textcolor{gray}{---} \\
\midrule
$-$ \textsc{Text}
& \cellcolor{red!12}$\times$ & $\checkmark$ & $\checkmark$ & $\checkmark$ & $\checkmark$
& $51.9\%$ & \textcolor{red!80!black}{$\downarrow\,13.1$} \\
$-$ \textsc{Visual}
& $\checkmark$ & \cellcolor{red!12}$\times$ & $\checkmark$ & $\checkmark$ & $\checkmark$
& $47.8\%$ & \cellcolor{red!12}\textcolor{red!80!black}{$\downarrow\,17.2$} \\
$-$ \textsc{Query}
& $\checkmark$ & $\checkmark$ & \cellcolor{red!12}$\times$ & $\checkmark$ & $\checkmark$
& $48.8\%$ & \textcolor{red!80!black}{$\downarrow\,16.2$} \\
$-$ \textsc{Structure}
& $\checkmark$ & $\checkmark$ & $\checkmark$ & \cellcolor{red!12}$\times$ & $\checkmark$
& $57.8\%$ & \textcolor{red!80!black}{$\downarrow\,7.2$} \\
$-$ \textsc{Spatial}
& $\checkmark$ & $\checkmark$ & $\checkmark$ & $\checkmark$ & \cellcolor{red!12}$\times$
& $61.1\%$ & \textcolor{red!80!black}{$\downarrow\,3.9$} \\
\bottomrule
\end{tabular}%
}
\vspace{-8pt}
\end{wraptable}

\textbf{Structure layout recovery.}
This tool recovers tables, forms, hierarchical text structure, and layout information. When needed, it also associates table cells with visual properties such as text color or background color. It is useful when the answer depends on both text and structure.

\textbf{Submit answer.}
This action terminates the evidence acquisition loop. The agent submits a final answer, short reasoning, and confidence level. The output is either an evidence-based answer or \texttt{Unknown} when the evidence is insufficient.

Not all five tools contribute equally. A leave-one-out ablation (Table~\ref{tab:tool_ablation}; analyzed in Section~\ref{sec:ablation}) removes each tool family in turn and measures the accuracy drop on DocVQA2026: Visual Inspection and Targeted Query are the most impactful, confirming that question-conditioned extraction---not the sheer number of tools---is what drives performance.

For long documents, we use a simple overlapping chunking strategy during evaluation. Each chunk $j$ is processed by the same Q-Guide loop and returns an answer $a_j$ with confidence $\sigma_j$; a lightweight adjudicator then combines the chunk-level answers by confidence-weighted majority,
\begin{equation}
    \hat{a} \;=\; \arg\max_{v}\;\sum_{j:\,a_j=v} w(\sigma_j),
    \qquad w(\textsc{low},\textsc{med},\textsc{high})=(1,2,3),
    \label{eq:fusion}
\end{equation}
falling back to an LLM arbitration over the candidate set only when~\eqref{eq:fusion} is tied. This proved more reliable than page-level retrieval for DocVQA2026, where many answers depend on nearby consecutive pages or cross-page context (chunking $65.0\%$ vs.\ page-level retrieval $50.0\%$; Section~\ref{sec:failure_discussion}). A question-aware retriever that selects relevant pages under a holistic view of the document is a natural extension we leave to future work.

\begin{figure*}[!htbp]
\centering
\includegraphics[width=0.92\textwidth]{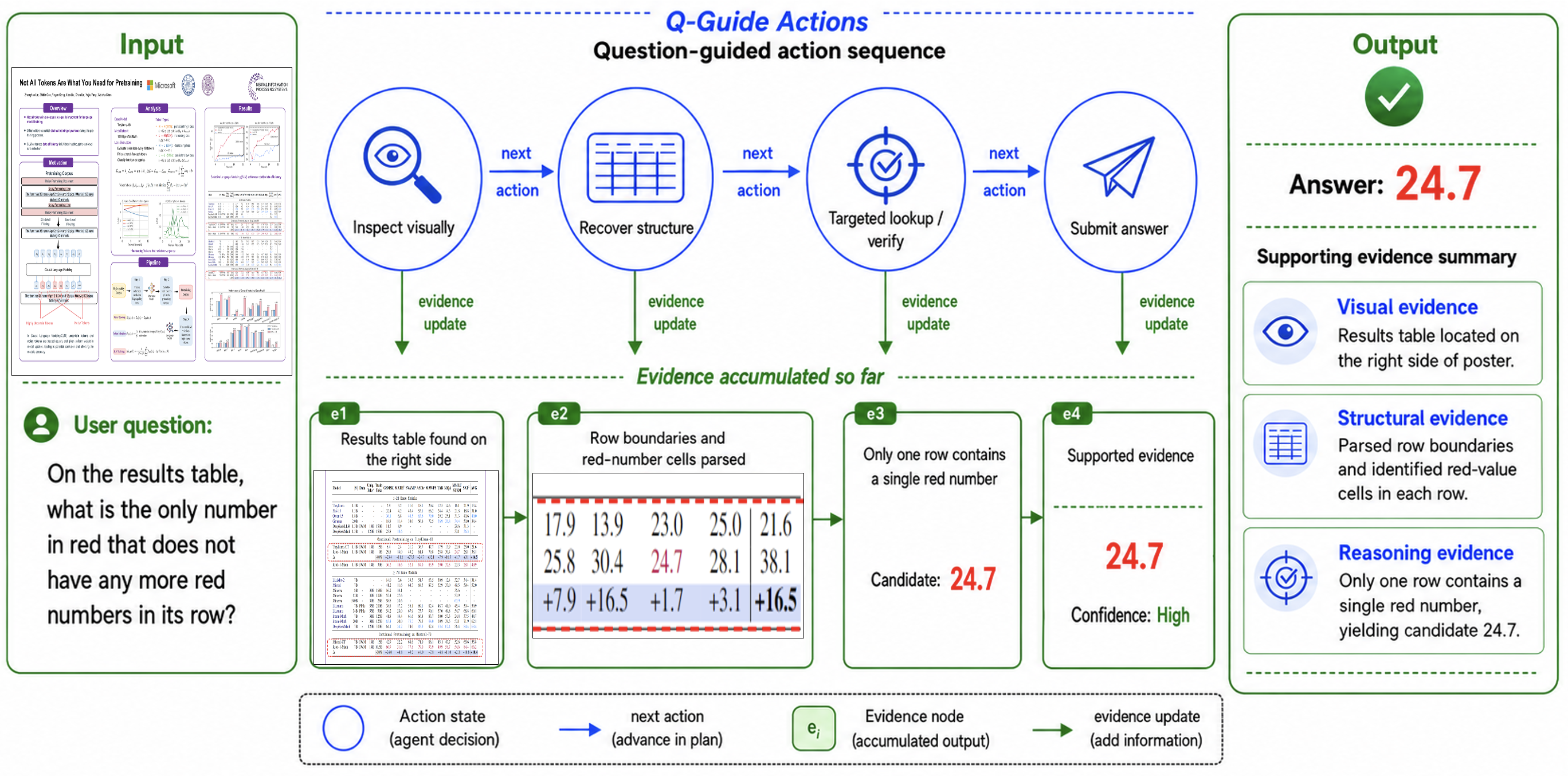}
\vspace{-2mm}
\caption{\textbf{Q-Guide in action.} A science-poster example where the answer depends on both table structure and color. The agent locates the relevant region, recovers the table and red cell values, and submits the answer only after the evidence supports it. Each evidence node $e_i$ shows the observation $o_i$ added at that step (Section~\ref{sec:formulation}); together they form the accumulated evidence state.}
\label{fig:qguide_sample}
\end{figure*}

\section{Implementation Details}
\label{sec:implementation}

This section grounds the three parts from Section~\ref{sec:architecture} in concrete components. The Q-Guide agent is a single MLLM (the \emph{LLM backbone} below); the question-conditioned recovery tools are backed by off-the-shelf perception services (\emph{tool backends}); and the self-assessment loop, including the low-confidence verification check, is realized as an explicit state machine (\emph{agent execution}).

\textbf{LLM backbone.}
We use Claude Opus 4.6~\cite{anthropicclaudeopus46} as the multimodal reasoning backbone. The model receives unstructured multimodal content and the user question, decides which tools to call, and produces the final answer. We use deterministic decoding with temperature 0. Page images are resized to preserve readability while fitting within the context budget.

\textbf{Tool backends.}
Structured text extraction, layout recovery, table/form parsing, and targeted region queries are implemented with Textract~\cite{amazontextract}. Spatial text detection uses a coordinate-aware text detector at word polygonal representation level~\cite{suryaocr}. For visual text grounding, i.e.\ anchoring detected text elements to their spatial coordinates in the document, we use Qwen3.5-2B~\cite{qwen3.5}, which provides bounding-box-level localization of textual elements. Visual inspection returns image crops from the source document. Color sampling and corridor text reading are implemented with lightweight image-processing routines. Region crops include padding and are upscaled before OCR or query-based extraction.

\textbf{Agent execution.}
The agent is implemented as a state machine using LangGraph~\cite{langgraph2024}, with nodes for initialization, agent reasoning, tool execution, answer extraction, low-confidence verification, and forced termination; this cleanly separates decision logic from tool execution and supports reproducible multi-turn interactions. The evidence-gathering budget is $T_{\max}=3$ turns, and multiple parallel tool calls in a single turn count as one turn. For long documents, we combine outputs from overlapping chunks via the confidence-weighted adjudication of Eq.~\ref{eq:fusion} (majority vote, with LLM arbitration on ties).

\begin{table*}[!htbp]
\centering
\small
\setlength{\tabcolsep}{4.5pt}
\renewcommand{\arraystretch}{1.15}
\resizebox{\textwidth}{!}{%
\begin{tabular}{@{}llcccccccc|c@{}}
\toprule
& & \multicolumn{8}{c}{\textbf{Document Category}} & \\
\cmidrule(lr){3-10}
\textbf{Method} & \textbf{Type} &
\shortstack{\textbf{Bus.}\\\textbf{Report}} &
\shortstack{\textbf{Comics}\\\phantom{x}} &
\shortstack{\textbf{Eng.}\\\textbf{Draw.}} &
\shortstack{\textbf{Infogr.}\\\phantom{x}} &
\shortstack{\textbf{Maps}\\\phantom{x}} &
\shortstack{\textbf{Sci.}\\\textbf{Paper}} &
\shortstack{\textbf{Sci.}\\\textbf{Poster}} &
\shortstack{\textbf{Slides}\\\phantom{x}} &
\textbf{Overall} \\
\midrule
\rowcolor{gray!8}
\multicolumn{11}{@{}l}{\textit{Direct prompting baselines (single LLM call, no tools)}} \\
\addlinespace[2pt]
Visual Prompt & Direct & $20.0\%$ & $40.0\%$ & $20.0\%$ & $20.0\%$ & $20.0\%$ & $30.0\%$ & $40.0\%$ & $30.0\%$ & $27.5\%$ \\
OCR Prompt & Direct & $30.0\%$ & $20.0\%$ & $10.0\%$ & $40.0\%$ & $50.0\%$ & $20.0\%$ & $50.0\%$ & $40.0\%$ & $32.5\%$ \\
Visual+OCR Prompt & Direct & $40.0\%$ & $20.0\%$ & $20.0\%$ & $50.0\%$ & $30.0\%$ & $50.0\%$ & $50.0\%$ & $50.0\%$ & $38.8\%$ \\
\midrule
\rowcolor{gray!8}
\multicolumn{11}{@{}l}{\textit{Agentic baselines (multi-step, tool use)}} \\
\addlinespace[2pt]
ARIAL~\cite{mohammadshirazi2025arial} & Agentic & $20.0\%$ & $40.0\%$ & $40.0\%$ & $30.0\%$ & $10.0\%$ & $30.0\%$ & $30.0\%$ & $60.0\%$ & $32.5\%$ \\
DocAgent~\cite{sun2025docagent} & Agentic & $30.0\%$ & $40.0\%$ & $40.0\%$ & $30.0\%$ & $30.0\%$ & $40.0\%$ & $50.0\%$ & $60.0\%$ & $40.0\%$ \\
MDocAgent~\cite{han2025mdocagent} & Agentic & $30.0\%$ & $40.0\%$ & $30.0\%$ & $40.0\%$ & $40.0\%$ & $50.0\%$ & $60.0\%$ & $30.0\%$ & $40.0\%$ \\
\midrule
\textbf{Q-Guide} & \textbf{Agentic} & $\mathbf{60.0\%}$ & $\mathbf{60.0\%}$ & $\mathbf{90.0\%}$ & $\mathbf{60.0\%}$ & $\mathbf{50.0\%}$ & $\mathbf{50.0\%}$ & $\mathbf{90.0\%}$ & $\mathbf{60.0\%}$ & $\mathbf{65.0\%}$ \\
\bottomrule
\end{tabular}%
}
\caption{\textbf{Results on the DocVQA2026 validation split} (80 questions, 8 categories). Q-Guide achieves the best overall accuracy, with the largest gains on engineering drawings and science posters, where precise local evidence is critical. Even at this sample size, Q-Guide's Wilson~\cite{wilson1927} $95\%$ confidence interval on the overall column ($[54.1,74.5]$) sits clearly above those of the strongest baselines (Visual+OCR $[28.8,49.7]$, DocAgent/MDocAgent $[30.0,51.0]$); per-category cells ($n=10$) are noted for trends only, and the gaps are robust under paired~\cite{mcnemar1947,dietterich1998} and bootstrap~\cite{efron1993bootstrap} tests.}
\label{tbl:docvqa_results}
\vspace{-8mm}
\end{table*}

\section{Experiments}
\label{sec:experiments}

\subsection{Datasets and Task Formulation}
\label{sec:dataset}

We evaluate Q-Guide on two deliberately complementary settings that exercise different failure modes of the same framework: DocVQA2026 stresses \emph{breadth} across document types and layouts, while M109NC stresses \emph{depth on one hard axis}---naming a character, which forces cross-page identity matching from low-resource Japanese text and visual appearance. DocVQA2026~\cite{docvqa2026dataset} contains $80$ questions across $8$ document categories: business reports, comics, engineering drawings, infographics, maps, science papers, science posters, and slides. Documents range from $1$ to $181$ pages, and many questions require correlating entities across regions, matching labels to visual elements, reading table cells, and combining evidence from multiple pages. Although compact, DocVQA2026 is deliberately adversarial rather than easy: the eight categories mix dense tables, engineering dimensions, colored cells, and map topology, stressing a breadth of perception that far larger single-domain benchmarks do not probe. Its difficulty is evident in the numbers---the strongest direct-prompting baseline reaches only $38.8\%$, and every prior agentic system we test stays at or below $40.0\%$---so headroom here reflects genuine perceptual challenge, not annotation noise.

The second setting is a character naming benchmark that we construct from Manga109~\cite{baek2026mangav26,multimedia_aizawa_2020}, inspired by the multi-task comic understanding setup of CoMiX~\cite{vivoli2024comix}. We call this task M109NC, for Manga109 Naming Characters. Given a set of context pages containing character names and a query page with a highlighted character (Figure~\ref{fig:manga109_sample}), the system must produce the character's canonical name. This requires active evidence acquisition beyond standard OCR: the model must read Japanese text from context pages, associate names with character appearances, and match the highlighted character on a different page. We use Manga109-v2026 annotations, which provide character body and face boxes linked to a book-level character roster.

For M109NC, we sample questions from $10$ Manga109 books selected for diversity and difficulty, yielding $504$ questions in total (roughly $50$ per book). Q-Guide is evaluated zero-shot, so no training on the remaining books is involved. For each book, we select a compact set of context pages via a greedy set-cover over character name mentions, then draw query samples from the non-context pages. We deliberately withhold the full book: the goal is to test whether the method can recover a character's identity from a small evidence set where names and visual references are present but not trivially matched.

\begin{wrapfigure}{r}{0.40\textwidth}
\vspace{-14pt}
\centering
\includegraphics[width=0.38\textwidth]{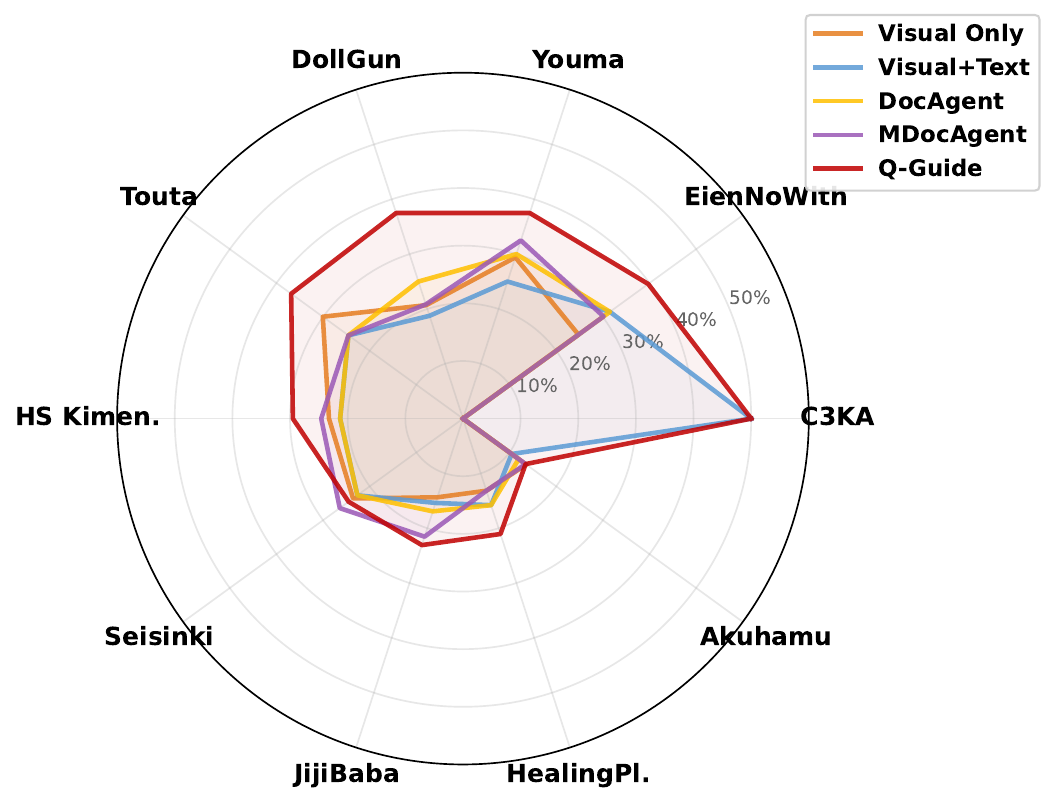}
\caption{\textbf{Per-book accuracy on M109NC.} Q-Guide (red) covers the largest area, indicating consistent gains across diverse manga styles.}
\label{fig:manga109_radar}
\vspace{-10pt}
\end{wrapfigure}
Performance varies across volumes (Figure~\ref{fig:manga109_radar}), tracking how visually distinctive and unambiguously named the characters are: Q-Guide is strongest on YoumaKourin and EienNoWith (unique hairstyles and clothing) and all methods struggle on Akuhamu, where many hamster characters share the same body shape.
For both benchmarks, we use ANLS-based matching as the primary evaluation criterion, supplemented by strict matching for numeric and date answers and order-invariant matching for list answers on DocVQA2026. Unless otherwise stated, all methods use Claude Opus 4.6 as a fixed reasoning-capable backbone LLM so that the comparison isolates the evidence-acquisition strategy from differences in the underlying model. We verify that Q-Guide's gains generalize across backbones; results with Claude Sonnet 4.6~\cite{anthropicclaudesonnet46} and Claude Opus 4.5~\cite{anthropicclaudeopus45} are reported in Section~\ref{sec:backbone}.

\subsection{Main Results}
\label{sec:main_results}

Table~\ref{tbl:docvqa_results} compares Q-Guide with direct prompting baselines (a single LLM call over page images, OCR text, or both) and recent agentic document-understanding frameworks on the DocVQA2026 validation split.


On DocVQA2026, Q-Guide reaches $65.0\%$, improving over the strongest direct baseline (Visual+OCR Prompt) by $26.2$ points and over DocAgent and MDocAgent by $25.0$ points---so giving the model both images and OCR text is not enough. These agentic baselines are recent multi-step frameworks: DocAgent~\cite{sun2025docagent} decomposes questions with a planning agent, while MDocAgent~\cite{han2025mdocagent} uses multi-agent reader/synthesizer collaboration. The largest gains appear on engineering drawings and science posters ($90.0\%$), which hinge on precise local evidence---dimensions, table cells, colored values, chart entries---where targeted lookup, visual inspection, and structure recovery beat global document processing. Q-Guide also improves on maps ($50.0\%$), though these remain challenging because many require spatial or topological reasoning that is only partially captured.

\begin{wraptable}{r}{0.48\textwidth}
\vspace{-24pt}
\centering
\small
\setlength{\tabcolsep}{4pt}
\renewcommand{\arraystretch}{1.08}
\caption{\textbf{Character naming on M109NC} ($n=504$). Q-Guide benefits strongly from higher-quality text: ground-truth OCR lifts its accuracy to $53.7\%$. The two Wilson $95\%$ intervals---$[28.4,36.5]$ under Claude-OCR and $[49.4,58.1]$ under GT-OCR---do not overlap, so this text-quality effect is well separated from sampling noise.}
\label{tab:manga109_results}
\scalebox{0.82}{%
\begin{tabular}{@{}lcc@{}}
\toprule
& \multicolumn{2}{c}{\textbf{Text Source}} \\
\cmidrule(lr){2-3}
\textbf{Method}
& \textbf{Claude-OCR}
& \textbf{GT-OCR} \\
\midrule
\rowcolor{gray!8}
\multicolumn{3}{@{}l}{\textit{Direct prompting baselines}} \\
\addlinespace[2pt]
Visual Prompt
& $21.8\%$
& $21.8\%$ \\
Visual+OCR Prompt
& $23.1\%$
& $24.9\%$ \\
\midrule
\rowcolor{gray!8}
\multicolumn{3}{@{}l}{\textit{Agentic approaches}} \\
\addlinespace[2pt]
DocAgent~\cite{sun2025docagent}
& $24.4\%$
& $29.3\%$ \\
MDocAgent~\cite{han2025mdocagent}
& $25.6\%$
& $27.8\%$ \\
\textbf{Q-Guide}
& $\mathbf{32.4\%}$
& $\mathbf{53.7\%}$ \\
\bottomrule
\end{tabular}%
}
\vspace{-8pt}
\end{wraptable}

Figure~\ref{fig:qguide_sample} illustrates the process on a science-poster example: the agent locates the relevant table, recovers its structure and color, then verifies the answer before submitting.
On M109NC, Q-Guide also improves over direct and agentic baselines (Table~\ref{tab:manga109_results}). With Claude-based OCR it reaches $32.4\%$, improving over Visual+OCR Prompt ($23.1\%$) by $9.3$ points and over DocAgent ($24.4\%$) by $8.0$ points; with ground-truth OCR it reaches $53.7\%$, a $28.8$-point gain over Visual+OCR Prompt with GT-OCR ($24.9\%$). Q-Guide's tools are thus highly effective when text is reliable, while imperfect OCR limits its potential. Notably, Q-Guide with imperfect Claude-OCR ($32.4\%$) already surpasses Visual+OCR Prompt with \emph{perfect} GT-OCR ($24.9\%$), suggesting that iterative tool use and cross-page visual matching add value beyond better text extraction alone.
One pattern holds across both benchmarks: visual input alone consistently trails structured multi-modal evidence (visual + text)---MLLMs reason better when evidence arrives already separated into complementary modalities than when they must decompose a scene internally. On M109NC this is especially striking, since the OCR is itself produced by Claude: the same model reads better when its own text extraction is surfaced as an explicit tool than end-to-end from pixels.

\subsection{Ablation Studies}
\label{sec:ablation}

Our central claim is that closing the perception gap comes from pointing perception at the right place, not from heavier reasoning control. We ablate two axes this implies: orchestration strategy (does adding planning, routing, or extra stages help?) and tool composition (which recovery modalities surface the missing evidence?).

\begin{figure*}[!htbp]
\centering
\begin{minipage}[c]{0.48\textwidth}
\centering
\scalebox{0.65}{%
\setlength{\tabcolsep}{3pt}
\renewcommand{\arraystretch}{1.08}
\begin{tabular}{@{}llcc@{}}
\toprule
\textbf{Configuration} & \textbf{Strategy} & \textbf{Acc.} & \textbf{$\Delta$} \\
\midrule
\rowcolor{gray!8}
\multicolumn{4}{@{}l}{\textit{Direct prompting baselines}} \\
\addlinespace[2pt]
Visual Prompt & Direct Answer & $27.5\%$ & --- \\
OCR Prompt & Direct Answer & $32.5\%$ & $+5.0$ \\
Visual+OCR Prompt & Direct Answer & $38.8\%$ & $+11.3$ \\
\midrule
\rowcolor{gray!8}
\multicolumn{4}{@{}l}{\textit{Two-stage agentic: research for the answer}} \\
\addlinespace[2pt]
Extract-Synthesize & Investigate $\to$ Answer & $44.3\%$ & $+16.8$ \\
\midrule
\rowcolor{gray!8}
\multicolumn{4}{@{}l}{\textit{Complex reasoning agentic}} \\
\addlinespace[2pt]
3-Stage Research & Plan $\to$ Gather $\to$ Answer & $35.7\%$ & $+8.2$ \\
Advisory Routing & Category-aware routing & $44.3\%$ & $+16.8$ \\
\midrule
\rowcolor{gray!8}
\multicolumn{4}{@{}l}{\textit{Tool-focused agentic}} \\
\addlinespace[2pt]
Basic agentic & Inspect $\circlearrowright$ OCR loop & $50.0\%$ & $+22.5$ \\
\textbf{Q-Guide} & \textbf{5 recovery tools} & $\mathbf{65.0\%}$ & $\mathbf{+37.5}$ \\
\bottomrule
\end{tabular}%
}
\end{minipage}%
\hfill
\begin{minipage}[c]{0.46\textwidth}
\centering
\includegraphics[width=\textwidth]{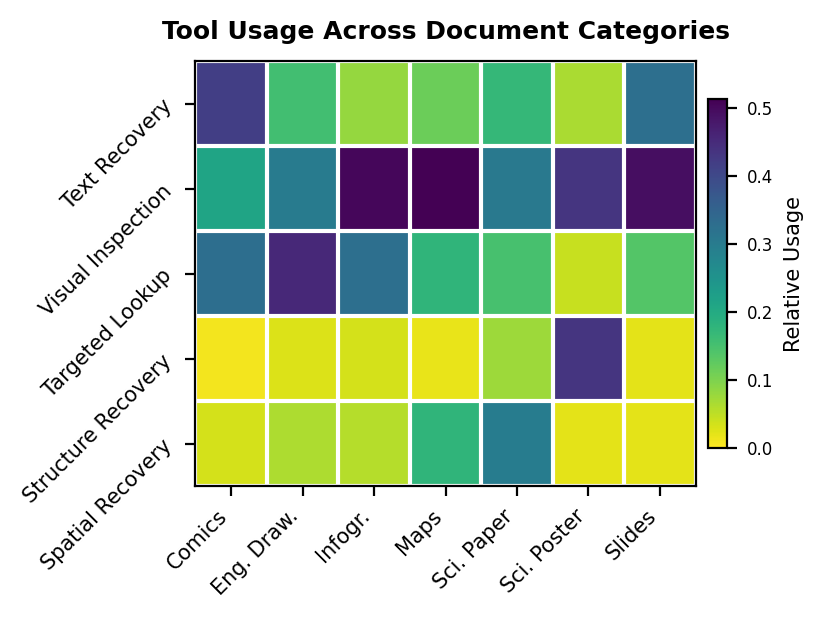}
\end{minipage}
\caption{\textbf{Orchestration and tool-use analysis.} \textit{Left}: Q-Guide performs best with a compact iterative tool-use loop, while heavier planning and routing do not improve accuracy. \textit{Right}: tool usage changes naturally across document categories, showing that the agent adapts evidence acquisition without explicit category routing.}
\label{fig:ablation_combined}
\end{figure*}

\textbf{Orchestration strategy.} Figure~\ref{fig:ablation_combined} (\textit{Left}) compares different ways of organizing the reasoning process. Direct prompting baselines range from $27.5\%$ to $38.8\%$, with Visual+OCR Prompt performing best among them. A simple two-stage agentic setup, Extract-Synthesize, reaches $44.3\%$, showing that separating investigation from answering already helps.

Yet more structure does not always help. The 3-stage research pipeline reaches only $35.7\%$---below the Visual+OCR baseline---and Advisory Routing ($44.3\%$) merely matches Extract-Synthesize: the bottleneck is not the absence of a plan. Extra planning produces brittle intermediate decisions, and category routing forces a fixed strategy even when the question needs different evidence. The strongest gains come from a simple iterative loop: Basic agentic reasoning reaches $50.0\%$ and Q-Guide $65.0\%$, a $37.5$-point improvement over Visual Prompt---the payoff is in perceptual recovery tools, not orchestration complexity. What that extra compute buys is visible in how accuracy grows with the perception budget (Fig.~\ref{fig:token_cost}, right): from $40.0\%$ at $T_{\max}=1$ to $55.0\%$ at $T_{\max}=2$ and $\sim\!65\%$ at $T_{\max}\geq3$, after which it plateaus---most of the gain is already captured within two to three deliberate rounds.

\begin{table}[!htbp]
\centering
\small
\scalebox{0.80}{%
\setlength{\tabcolsep}{5pt}
\renewcommand{\arraystretch}{1.12}
\begin{tabular}{@{}lccc|ccc@{}}
\toprule
& \multicolumn{3}{c|}{\textbf{DocVQA2026}} & \multicolumn{3}{c}{\textbf{M109NC}} \\
\cmidrule(lr){2-4}\cmidrule(lr){5-7}
\textbf{Method}
& \textbf{Opus 4.6}
& \textbf{Sonnet 4.6}
& \textbf{Opus 4.5}
& \textbf{Opus 4.6}
& \textbf{Sonnet 4.6}
& \textbf{Opus 4.5} \\
\midrule
\rowcolor{gray!8}
\multicolumn{7}{@{}l}{\textit{Direct prompting baselines}} \\
\addlinespace[2pt]
Visual Prompt & $27.5\%$ & $30.0\%$ & $27.1\%$ & $21.8\%$ & $21.8\%$ & $21.8\%$ \\
Visual+OCR Prompt & $38.8\%$ & $31.4\%$ & $37.1\%$ & $23.1\%$ & $22.9\%$ & $22.9\%$ \\
\midrule
\rowcolor{gray!8}
\multicolumn{7}{@{}l}{\textit{Agentic baselines}} \\
\addlinespace[2pt]
DocAgent & $40.0\%$ & $34.3\%$ & $37.1\%$ & $24.4\%$ & $24.2\%$ & $22.5\%$ \\
MDocAgent & $40.0\%$ & $31.4\%$ & $31.4\%$ & $25.6\%$ & $23.8\%$ & $25.1\%$ \\
\midrule
\textbf{Q-Guide} & $\mathbf{65.0\%}$ & $\mathbf{62.8\%}$ & $\mathbf{64.2\%}$ & $\mathbf{32.4\%}$ & $\mathbf{32.2\%}$ & $\mathbf{31.5\%}$ \\
\bottomrule
\end{tabular}%
}
\caption{\textbf{Accuracy across MLLM backbones.} Q-Guide consistently achieves the best accuracy regardless of the underlying model. On M109NC, direct prompting baselines plateau around $22\%$ irrespective of backbone, suggesting the bottleneck is the reasoning architecture rather than perceptual capability.}
\label{tab:backbone}
\end{table}

\textbf{Tool composition.} Figure~\ref{fig:ablation_combined} (\textit{Right}) shows the agent is not a fixed pipeline: it reaches for text recovery on text-heavy documents, visual inspection when local detail matters, and structure or spatial recovery for specialized cases---the question directs perception toward the relevant modality gap. Removing tools one at a time (Table~\ref{tab:tool_ablation}) shows how much each one carries: dropping Visual Inspection or Targeted Query costs the most ($-17.2$ and $-16.2$ points), Text Recovery somewhat less ($-13.1$), and Structure and Spatial least ($-7.2$ and $-3.9$). Tellingly, losing just those top two together ($-33.4$ points) wipes out more than Q-Guide's entire lead over the best baseline---so it is the question-driven \emph{targeting} of evidence, not the length of the toolset, that does the work.

\begin{figure}[!htbp]
\centering
\includegraphics[width=0.95\textwidth]{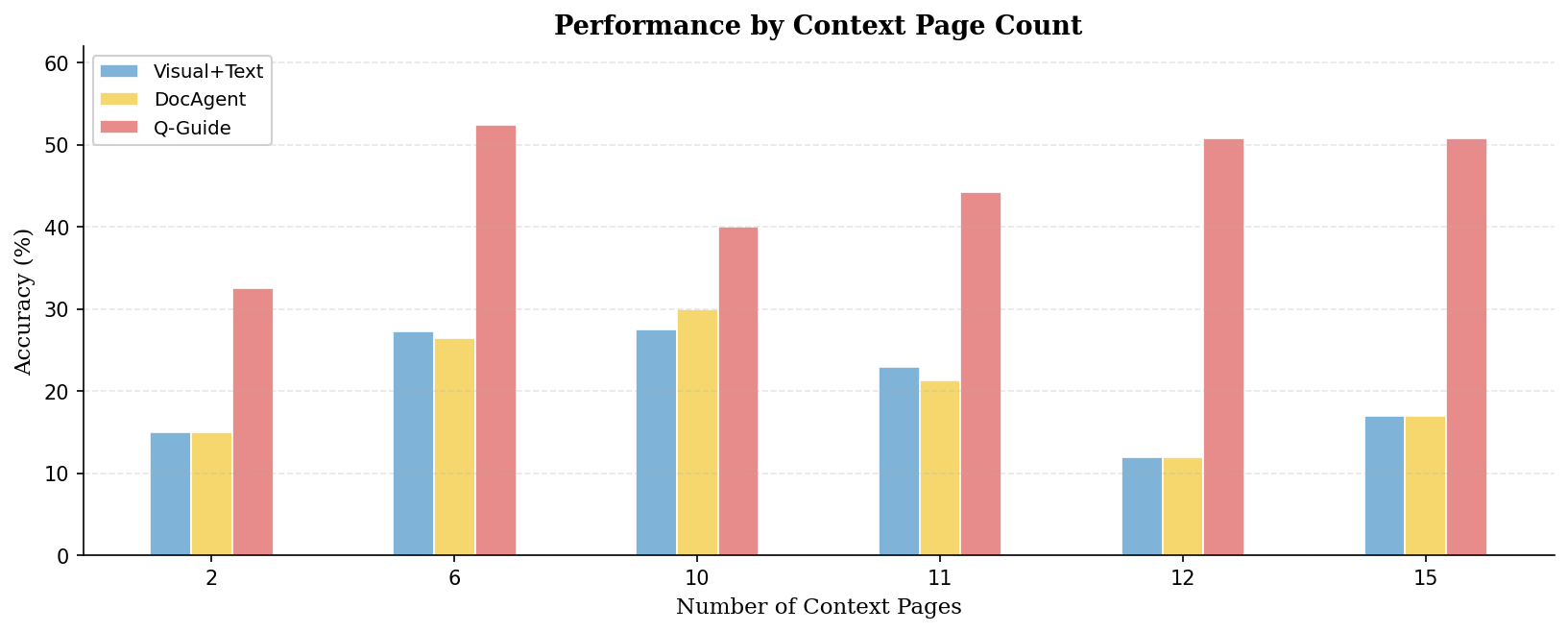}
\caption{\textbf{M109NC accuracy by context page count.} Q-Guide leads at every context size and exploits additional naming pages that single-call baselines cannot, exceeding $50\%$ with $12$ or more naming pages while the baselines stay nearly flat.}
\label{fig:context_bars}
\end{figure}

On M109NC, the $+21.3$-point Claude-OCR-to-GT-OCR gap ($32.4\%\!\to\!53.7\%$) shows the payoff is largest when recovered text is reliable, yet Q-Guide still beats all baselines under imperfect OCR. Broken down by context-page count (Figure~\ref{fig:context_bars}), Q-Guide leads at every context size and scales with available evidence, exceeding $50\%$ with $12$ or more naming pages, whereas the baselines stay nearly flat.

\subsection{LLM Backbone Versatility}
\label{sec:backbone}

To check that Q-Guide's gains generalize beyond one backbone, we re-run all methods with Claude Sonnet 4.6~\cite{anthropicclaudesonnet46} and Claude Opus 4.5~\cite{anthropicclaudeopus45}; Table~\ref{tab:backbone} reports the results. Q-Guide improves over every baseline regardless of the model, so the gains come from the evidence-acquisition strategy rather than a specific model's capability. Stronger backbones amplify its DocVQA2026 advantage, but weaker ones still recover substantially with its tools. On M109NC the direct baselines are flat across models ($\sim$22\%) while Q-Guide holds $\sim$32\%, so character naming benefits more from structured evidence acquisition than from raw perceptual capability. MDocAgent, by contrast, degrades on DocVQA2026 with weaker backbones ($40.0\%\!\to\!31.4\%$): its multi-agent orchestration is fragile under reduced capability, while Q-Guide's simpler loop stays robust.

\subsection{Limitations}
\label{sec:failure_discussion}

Q-Guide's gains come with limitations that also mark where the method can improve: it costs more compute than single-pass prompting, it still struggles with spatial and topological content, and it remains bounded by the quality of the text it recovers.

\begin{figure}[!htbp]
\centering
\begin{minipage}[c]{0.48\textwidth}
\centering
\includegraphics[width=\textwidth, height=0.62\textwidth, keepaspectratio=false]{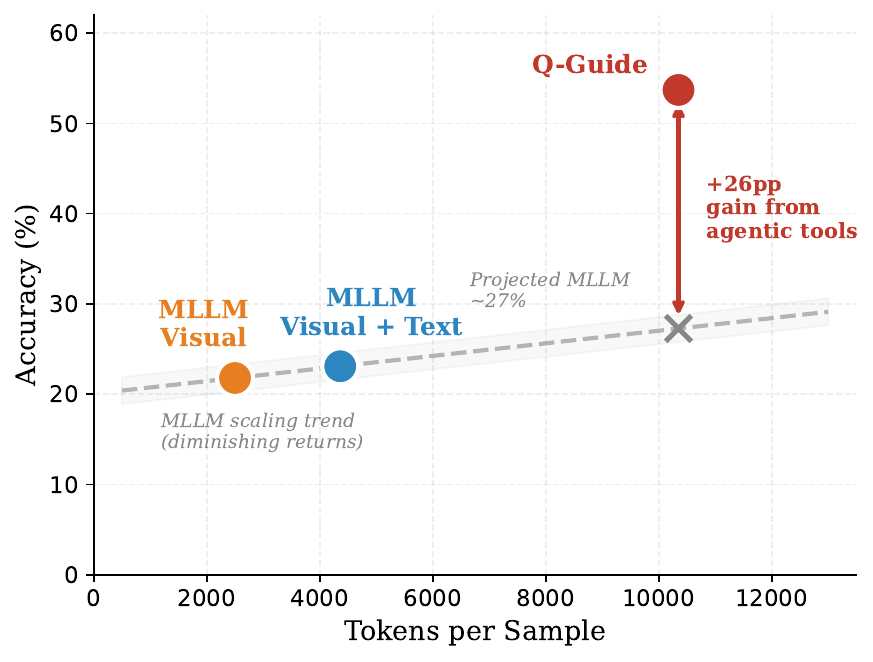}
\end{minipage}%
\hfill
\begin{minipage}[c]{0.48\textwidth}
\centering
\includegraphics[width=\textwidth, height=0.62\textwidth, keepaspectratio=false]{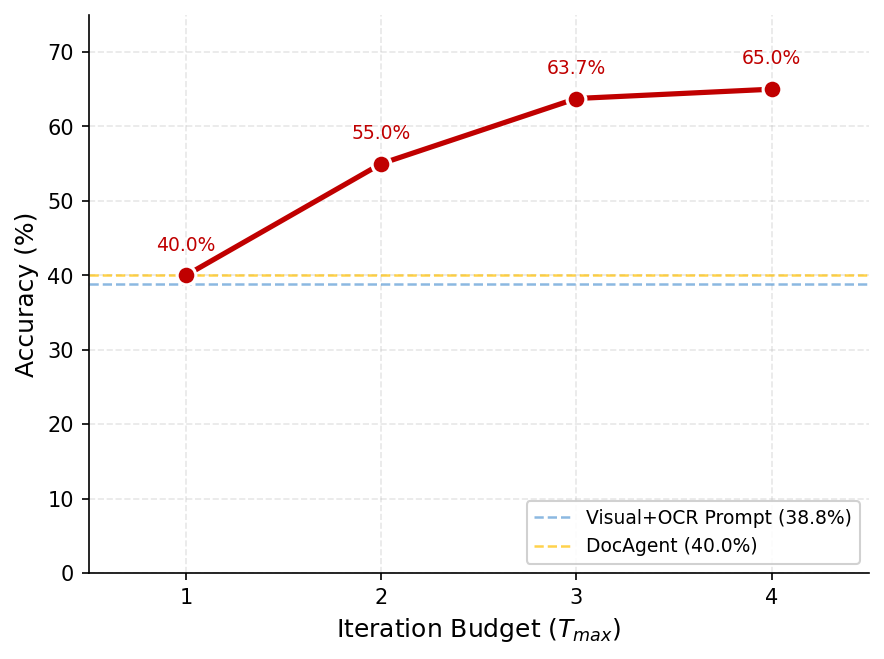}
\end{minipage}
\caption{\textbf{Cost and budget analysis.} \textit{Left:} accuracy vs.\ token cost on M109NC; Q-Guide breaks above the single-call MLLM ceiling (dashed, $+26$pp) by spending tokens on active investigation rather than context stuffing. \textit{Right:} accuracy vs.\ iteration budget ($T_{\max}$) on DocVQA2026; most gain is captured within 2--3 rounds, with diminishing returns beyond $T_{\max}=3$.}
\label{fig:token_cost}
\end{figure}

\textbf{Compute cost.} Token use follows the simple relation $\text{Tokens}\approx \alpha P_{\text{ctx}}+\beta\sum_i|y_i|+\gamma T$---one term per cost source: page images in context, recovered tool outputs, and reasoning turns---so cost scales with the rounds and evidence gathered, not with document length. Table~\ref{tab:tokens} bears this out on M109NC: Q-Guide's overhead is overwhelmingly input-side ($9{,}668$ of $10{,}351$ tokens), i.e.\ surfacing recovered evidence rather than longer generation, and it spends $\sim\!2.5\times$ the tokens of the strongest agentic baseline. Since this cost grows with $T$ while accuracy saturates by $T\!\approx\!3$ (Fig.~\ref{fig:token_cost}), a small budget captures most of the benefit, but the method is materially more expensive than a single pass.

\textbf{Spatial and topological reasoning.} Q-Guide's clearest accuracy weakness is on map questions, which often hinge on a value that depends on a route, adjacency, or orientation the model must follow visually. Its tools recover local text and detail but do not represent the map as a navigable structure, so a dedicated map-to-graph tool is the clearest next step. Relatedly, long-document handling relies on a simple fixed-window chunking with cross-chunk adjudication; this already beats page-level retrieval ($65.0\%$ vs.\ $50.0\%$, as many answers span consecutive pages), but a question-aware retriever remains future work.

\begin{wraptable}{r}{0.44\textwidth}
\vspace{-14pt}
\centering
\small
\setlength{\tabcolsep}{6pt}
\renewcommand{\arraystretch}{1.12}
\caption{\textbf{Measured per-question token cost on M109NC} (mean over $50$ questions, Claude Opus 4.6). Q-Guide's overhead is input-dominated ($93\%$).}
\label{tab:tokens}
\begin{tabular}{@{}lccc@{}}
\toprule
\textbf{Method} & \textbf{In} & \textbf{Out} & \textbf{Total} \\
\midrule
Visual Only   & $2{,}336$ & $166$ & $2{,}502$ \\
Visual+Text   & $3{,}941$ & $422$ & $4{,}364$ \\
DocAgent      & $3{,}452$ & $603$ & $4{,}055$ \\
\textbf{Q-Guide} & $\mathbf{9{,}668}$ & $\mathbf{683}$ & $\mathbf{10{,}351}$ \\
\bottomrule
\end{tabular}
\vspace{-8pt}
\end{wraptable}

\textbf{OCR dependence.} Q-Guide is only as good as the text it can read. Accuracy on M109NC rises almost linearly with text quality---from $21.8\%$ with no text (image only), to $32.4\%$ with Claude's own OCR, to $53.7\%$ with ground-truth text---so if we place these on a quality scale from $0$ to $1$, the trend follows $\mathrm{acc}\!\approx\!21.8\%+31.9x$. On that scale Claude-OCR lands at only $x\!\approx\!0.33$: it realizes about a third of the gain that perfect text would, which tells us how much of the remaining error is bad OCR, and how much headroom better text extraction would unlock.

\textbf{Cross-page matching.} On M109NC the agent resolves character identity by combining visual-appearance similarity with the textual name mentions gathered across context pages, evaluating both natively from tool-surfaced evidence rather than through a separate face-embedding pipeline. Its residual errors are dominated by perceptual visual-matching failures rather than language-level coreference: single-call methods hit a ceiling regardless of context size, while Q-Guide breaks through by spending tokens on active tool use rather than passive input expansion (Fig.~\ref{fig:token_cost}).

\section{Conclusion}
\label{sec:conclusion}

Q-Guide is a compact agent that spends test-time compute deciding what to perceive, directing recovery tools toward the evidence a question needs. It outperforms both single-pass (System-1) prompting and heavier multi-agent orchestration across three backbones and two visually distinct domains, with accuracy scaling with the perception budget---most of the gain arriving within 2--3 rounds. For document VQA, slow thinking is thus most useful applied to perception itself, not only to language-level reasoning: across cost, the confidence gate, and residual failures alike, the lever for further progress is richer \emph{perception}, not deeper control logic. That the same loop carried unchanged from documents to manga character naming suggests a general recipe for evidence-limited multimodal QA; the clearest next steps are a structure-aware tool for maps, a question-aware retriever for long documents, and a port to open-weight backbones.

\clearpage
\bibliographystyle{splncs04}
\bibliography{library}

\end{document}